\documentclass[sigconf,camera-ready]{acmart}
\AtBeginDocument{%
  \providecommand\BibTeX{{%
    \normalfont B\kern-0.5em{\scshape i\kern-0.25em b}\kern-0.8em\TeX}}}

\setcopyright{acmcopyright}
\copyrightyear{2023}
\acmYear{2023}
\acmDOI{xxxx/xxxx}

\acmConference[KDD '23 workshop]{KDD '23 Workshop on Machine Learning in Finance}{August, 2023}{Workshop}
\acmBooktitle{KDD '23 Workshop on Machine Learning in Finance, August, 2023, Workshop}
\acmPrice{xx}
\acmISBN{xxx-x-xxxx-xxxx-X/xx/xx}

\usepackage{multirow}
\usepackage{bm}
\usepackage{algorithm}
\usepackage{algorithmic}
\usepackage{booktabs}
\usepackage{amsmath}
\usepackage{mathtools}
\usepackage{tikz}
\usepackage{xcolor}
\usetikzlibrary{arrows}
\usepackage{nicefrac} 

\usepackage{algorithm}
\usepackage{algorithmic}
\usepackage{bm}
\usepackage{subcaption}

\newcommand{\abs}[1]{\left|#1\right|}

\DeclareMathOperator*{\argmax}{arg\,max}

\begin{document}

\title{Incorporating Pre-trained Model Prompting in Multimodal \\ Stock Volume Movement Prediction}
\renewcommand{\shorttitle}{Incorporating Pre-trained Model Prompting in Multimodal Stock Volume Movement Prediction}

\author{Ruibo Chen$^*$}
\email{ruibochen@pku.edu.cn}
\affiliation{Peking University\country{China}}

\author{Zhiyuan Zhang$^*$}
\email{zzy1210@pku.edu.cn}
\affiliation{Peking University\country{China}}

\author{Yi Liu}
\email{yliu.pku@outlook.com}
\affiliation{Peking University\country{China}}

\author{Ruihan Bao}
\email{ruihan.bao@mizuho-sc.com}
\affiliation{Mizuho Securities Co., Ltd\country{Japan}}

\author{Keiko Harimoto}
\email{keiko.harimoto@mizuho-sc.com}
\affiliation{Mizuho Securities Co., Ltd\country{Japan}}

\author{Xu Sun}
\email{xusun@pku.edu.cn}
\affiliation{Peking University\country{China}}

\renewcommand{\shortauthors}{Ruibo Chen, et al.}

\begin{abstract}

Multimodal stock trading volume movement prediction with stock-related news is one of the fundamental problems in the financial area. Existing multimodal works that train models from scratch face the problem of lacking universal knowledge when modeling financial news. In addition, the models' ability may be limited by the lack of domain-related knowledge due to insufficient data in the datasets. To handle this issue, we propose the \textbf{Pro}mpt-based \textbf{MU}ltimodal \textbf{S}tock volum\textbf{E} prediction model (ProMUSE) to process text and time series modalities. We use pre-trained language models for better comprehension of financial news and adopt prompt learning methods to leverage their capability in universal knowledge to model textual information. Besides, simply fusing two modalities can cause harm to the unimodal representations. Thus, we propose a novel cross-modality contrastive alignment while reserving the unimodal heads beside the fusion head to mitigate this problem. Extensive experiments demonstrate that our proposed ProMUSE outperforms existing baselines. Comprehensive analyses further validate the effectiveness of our architecture compared to potential variants and learning mechanisms. Our code will be available in \href{https://github.com/RayRuiboChen/ProMUSE}{https://github.com/RayRuiboChen/ProMUSE}.

\end{abstract}



\maketitle

\def\thefootnote{*}\footnotetext{Equal contribution}\def\thefootnote{\arabic{footnote}}

\section{Introduction}

Stock trading volume movement prediction is one of the fundamental tasks in the financial area which has been paid much attention to~\citep{ajinkya1989behavior,cartea2016closed,liu2017intraday}, and it has various important downstream applications such as algorithmic trading~\citep{treleaven2013algorithmic,hendershott2013algorithmic} and stock trading anomaly detection~\citep{nelson1998time}.

Traditional researches only use historical trading data and rely heavily on feature engineering. They employ statistical models to forecast the time series like the Autoregressive Integrated Moving Average model (ARIMA)~\citep{box1970distribution,nelson1998time}. With the development of deep learning techniques, \citet{gharehchopogh2013linear} use linear regression for stock market trading volume prediction. More complicated models utilizing LSTM and CNN are also applied in this field~\citep{chen2015lstm,selvin2017stock,lu2021cnn}. However, these methods which only involve historical trading data suffer from the lack of basic stock information. Human traders make their decisions on multiple factors, including stock-related news. As a result, unimodal models using only historical trading data as the input may make incorrect predictions. Thus, researchers start to introduce text information like news and tweets to better model the stock movement. Typically, sentimental analysis modules are used to reflect the market sentiment~\citep{nguyen2015sentiment,li2014news,mittal2012stock,pagolu2016sentiment}. These methods usually adopt pipeline architecture and the two modalities are not integrated properly. Additional errors may get involved in the prediction of sentiment. More recent works design multimodal models to jointly process text and time series data in order to acquire a better understanding. For example, \citet{8966989} propose to use the event-driven LSTM model to leverage news data. \citet{zou2022multimodal} design a hybrid multimodal model using CNN and SVM as the backbone.

A significant weakness of previous works is that they tend to construct a certain architecture and train from scratch. Current high-quality financial news datasets all tend to be much smaller in size than the large-scale unlabelled corpora collected from the Internet since it is costly to write, collect and filter to get the related news. Thus, it is very difficult to train a robust large multimodal model based on them, causing inferior ability compared to pre-trained language models such as Fin-BERT~\citep{yang2020finbert} and ChatGPT\footnote{\href{https://openai.com/product/chatgpt}{https://openai.com/product/chatgpt}}. In addition, as the topics and contents of financial news can be broad, domain knowledge and universal knowledge are both required for the textual learning process, making incorporating pre-trained language models necessary in this task.

\begin{figure*}[!h]
    \centering
    \includegraphics{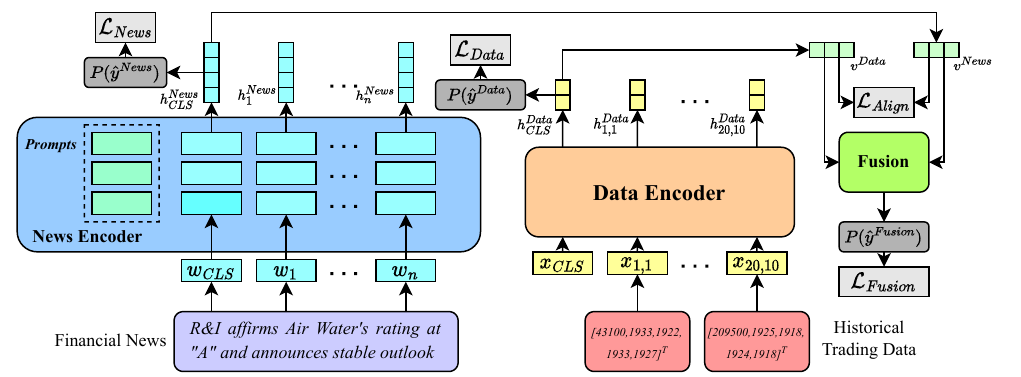}
    \caption{An overview of our model. Financial news and historical trading data are encoded respectively. The News Encoder employs a frozen Financial-RoBERTa as the backbone and only the continuous prompts in all layers are tunable. The Data Encoder employs a pre-trained 6-layer transformer which will be fine-tuned during training. The fusion module utilizes news and trading data representations to obtain an integrated prediction. The alignment loss is designed for cross-modality contrastive alignment and prevents damage to the unimodal representations during training.}
    \label{fig:model_architecture}
    \vskip -0.15 in
\end{figure*}

To settle the issue of lack of necessary knowledge, we propose the \textbf{Pro}mpt-based \textbf{MU}ltimodal \textbf{S}tock volum\textbf{E} movement prediction model (ProMUSE), which includes a News Encoder, a Data Encoder, and a fusion module. As illustrated in Figure~\ref{fig:model_architecture}, the News Encoder uses a pre-trained language model as a backbone, and we use prompt learning methods to efficiently exploit the knowledge within the pre-trained models. However, the direct fusion of the two encoders will severely damage the representation learning of the two modalities. Thus we propose to reserve a prediction head for each encoder to receive unimodal supervision. Specially, we add a cross-modality contrastive alignment loss to align the embedding space, alleviating the harm to their representations during the joint training. For inference, we use the algorithmic mean of the two unimodal head predictions and the multimodal fusion result through the fusion module. This makes it possible for our model to robustly generate outputs with even unimodal inputs.

We validate our proposed ProMUSE model on the TOPIX500 trading dataset and overnight financial news~\citep{li2021modeling} from Reuters. Our extensive experiments demonstrate that our proposed ProMUSE outperforms unimodal methods and multimodal baselines with significant gaps. We also conduct a series of ablation studies to analyze different modules in our model. We show that fusion-only methods perform worse as they lack unimodal supervision and therefore incur damage to unimodal representations. on the other hand, ensemble-only methods lose cross-modality contrastive alignment. Our method implements multimodal fusion and alignment reserving unimodal prediction heads to mitigate the harm to representation learning, thus achieving the best results. Besides, our proposed ProMUSE can reach higher performance than methods training from scratch consistently under varying data sizes. 

Our main contributions can be summarized as follows:
\begin{itemize}
    \item We introduce pre-trained language model prompting into multimodal stock volume movement prediction task to bring domain knowledge and universal knowledge with limited datasets. 
    \item We propose a multimodal method ProMUSE to incorporate knowledge from texts and historical trading data. Both unimodal and multimodal supervision training are used together with cross-modality contrastive alignment to alleviate the damage to the representation learning during multimodal learning.
    \item Experimental results show that our method significantly outperforms competitive baselines and further analyses validate the effectiveness of our proposed modules in ProMUSE.
\end{itemize}

\section{Methodology}

In this section, we first introduce the task formulation of multimodal stock volume movement prediction. Then we specify the architecture of our multimodal stock trading volume movement prediction model, the contrastive loss for cross-modality alignment, and present our training objectives and inference algorithms finally.

\subsection{Task Formulation}

The overnight stock volume movement prediction task is a binary classification problem: label 1/0 denotes that the volume goes up/down on the next trading day. The model's classification decisions are based on the overnight news and the stock historical trading data for the past 20 days, which both serve as the model's input.

\paragraph{Overnight News.}
Following~\citep{li2021modeling,zhao2021long}, we only adopt news headlines for our task, as they are more informative with a suitable length for processing. The overnight news can be modeled as an input sentence with $n$ tokens $W=\left\{w_1,w_2,\cdots,w_n\right\}$. The overnight news happened after the close of the $20^\text{th}$ day's trading market and before the opening of the next day.

\paragraph{Stock Historical Trading Data.} 

The overall stock historical trading data $X$ includes trading volumes and prices for the past 20 days and 10 time slots for each day with a granularity of 30 minutes. The trading data $x$ of a specific time slot includes the volume $x^v$, high price $x^h$, low price $x^l$, open price $x^o$, and close price $x^c$. The overall stock historical trading data $X$ can be formulated as $X=\{x_{i,j}|i\in[1,20]\cap \mathbb{Z}, j\in[1,10]\cap \mathbb{Z}\}$, where $x_{i,j}$ represents the trading data for the $j$-th time slot in the $i$-th day, namely $x_{i,j}=[x^v_{i,j}, x^h_{i,j}, x^l_{i,j}, x^o_{i,j}, x^c_{i,j}]^\text{T}\in\mathbb{R}^5$. 

~\\
Our goal is to predict the volume movement of the first time slot for the $21^\text{st}$ day $x_{21,1}^v$, which is the first 30 minutes after the opening of the market. Following \citet{zhao2021long}, we define the movement here as a comparison between $x_{21,1}^v$ and the average volume in the same time slot for the past 20 days, so the final prediction target label $y$ can be formulated as:
\begin{align}
\begin{split}
y=\mathbb{I}\left(x_{21,1}^v>\bar{v}\right)=\left\{
\begin{array}{lr}
1, &\text{if}\quad x_{21,1}^v > \bar{v}   \\
0, &\text{if}\quad x_{21,1}^v \le \bar{v}
\end{array}
\right.,
\end{split}
\end{align}
where $\bar{v}$ denotes the average volume in the first slot for the past 20 days and is defined as: 
$\bar{v}= \frac{1}{20}\sum_{i=1}^{20} x_{i,1}^v$.

Then our dataset can be denoted as $\mathcal{D}=\{(X^{(i)},W^{(i)},y^{(i)})\}_{i=1}^N$. We adopt the accuracy (ACC) metric to evaluate the performance:

\begin{equation}
    \text{ACC}=100\%\times\sum\limits_{i=1}^{N}\mathbb{I}\left(\argmax\limits_y P(y|X^{(i)},W^{(i)})=y^{(i)}\right),
\end{equation}
where $X$ denotes the input historical trading data, $W$ denotes the stock-related news, $y$ denotes the target label, $P(y|X^{(i)},W^{(i)})$ denotes the predicted probabilities, and $\argmax_y P(y|X^{(i)},W^{(i)})$ denotes the predicted label. 

\subsection{Model Architecture Overview}

Our proposed ProMUSE mainly includes two large-scale pre-trained modules to process the input: the News Encoder for the text modality, and the Data Encoder for the time series modality. The two encoders can effectively capture the features from the input and transform them into corresponding vectors. They also use their prediction head to produce unimodal losses and prediction to receive unimodal supervision, which will be introduced in section~\ref{sec:news_encoder} and section~\ref{sec:data_encoder}. In section~\ref{sec:fusion}, we further utilize a fusion module for generating multimodal losses and outputs. The cross-modality contrastive alignment described in section~\ref{sec:alignment} is designed to help improve unimodal representations during multimodal learning. Finally, we use the weighted sum of the two unimodal losses, multimodal loss, and alignment loss for the training objective in section~\ref{sec:training_objective}. During inference in section~\ref{sec:inference}, unimodal and multimodal predictions are combined. An overview of the model architecture is shown in Figure~\ref{fig:model_architecture}.

\subsection{News Encoder} 
\label{sec:news_encoder}

The News Encoder is Financial-RoBERTa\footnote{\href{https://huggingface.co/soleimanian/financial-roberta-large-sentiment}{https://huggingface.co/soleimanian/financial-roberta-large-sentiment}}, a 24-layer RoBERTa~\citep{roberta} model pre-trained on financial text, such as financial statements, news, and earnings announcements. To avoid overfitting on the limited news data as well as to accelerate the training procedure, we adopt prompt learning methods and choose P-Tuning v2~\citep{ptuningv2} as it reaches the best performance in our preliminary experiments. This method inserts soft prompts in each layer of the Transformer-based models. We set the prompt length to 20 and enable the reparameterization. In our experiments, only the prompts are tunable, while all other parameters in the language model are frozen. 

Given the input news $W$ including $n$ tokens, we insert a CLS token at the beginning of the sentence: $\left\{w_\text{CLS},w_1,w_2,\cdots,w_n\right\}$. Then we utilize the Financial-RoBERTa and P-Tuning v2 method to transform it into a series of vectors $\{h_\text{CLS}^\text{News},h_1^\text{News},h_2^\text{News},...,h_n^\text{News}\}$:
\begin{align}
\{h_\text{CLS}^\text{News},\cdots,h_n^\text{News}\} = \text{News-Encoder}\left(\left\{w_\text{CLS},\cdots,w_n\right\}\right),
\end{align}
where $h_i^\text{News} \in \mathbb{R}^{d_\text{News}}$, $d_\text{News}$ is the hidden size of the language model, and $d_\text{News}=1024$ in our settings. 

We further obtain a unimodal prediction $P\big(\hat{y}^\text{News}\big|h_\text{CLS}^\text{News}\big)$ according to $h_\text{CLS}^\text{News}$ by a linear classification head $\text{Linear-Head}^\text{News}$ to predict the scores of each categories, $\text{Linear-Head}^\text{News}\left(h_\text{CLS}^\text{News}\right) \in \mathbb{R}^2$:

\begin{equation}
P\big(\hat{y}^\text{News}\big|h_\text{CLS}^\text{News}\big)=\text{Softmax}\left( \text{Linear-Head}^\text{News}\left(h_\text{CLS}^\text{News}\right) \right).
\end{equation}

The unimodal loss for the News Encoder head is the Cross-entropy loss:
\begin{equation}
    \mathcal{L}_\text{News}=-\log P\big(\hat{y}^\text{News}=y\big|h_\text{CLS}^\text{News}\big),
\end{equation}

\subsection{Data Encoder} 
\label{sec:data_encoder}

In order to acquire high-quality representations for historical trading data, we pre-train a 6-layer Transformer~\citep{transformer} model. Recall that historical trading data $X=\{x_{i,j}|i\in[1,20]\cap \mathbb{Z}, j\in[1,10]\cap \mathbb{Z}\}$, $x_{i,j} \in \mathbb{R}^5$, the input time series can therefore be formulated as $[x_\text{CLS};x_{1,1};x_{1,2};x_{1,3};\cdots;x_{2,1};\cdots;x_{20,10}]\in\mathbb{R}^{201\times 5}$ in the time order, where the time slot number is $1+20\times 10=201$. The Data Encoder transforms the input series into a hidden vector series $[h_\text{CLS}^\text{Data};h_{1,1}^\text{Data};h_{1,2}^\text{Data};h_{1,3}^\text{Data};\cdots;h_{2,1}^\text{Data};\cdots;h_{20,10}^\text{Data}]\in\mathbb{R}^{201\times d_\text{Data}}$:
\begin{align}
[h_\text{CLS}^\text{Data};\cdots;h_{20,10}^\text{Data}]=\text{Data-Encoder}\big([x_\text{CLS};\cdots;x_{20,10}]\big),
\end{align}
where $h_{i,j}^\text{Data} \in \mathbb{R}^{d_\text{Data}}$. Here $d_\text{Data}$ is the hidden size of the Transformer, and we set $d_\text{Data}=200$. 

During the pre-training phase, we use a linear head after $h_\text{CLS}^\text{Data}$ to directly predict $x_{21,1}^v$, the volume for the first time slot in the $21^\text{st}$ day. We adopt Mean Square Error (MSE) loss for optimization:
\begin{equation}
    \mathcal{L}_\text{MSE}=(\text{Linear-Pre}(h_\text{CLS}^\text{Data})-x_{21,1}^v)^2.
\end{equation}

For multimodal stock volume movement prediction, similar to the News Encoder, we get the unimodal prediction $P\big(\hat{y}^\text{Data}\big|h_\text{CLS}^\text{Data}\big)$ and unimodal loss $\mathcal{L}_\text{Data}$ as follows:

\begin{align}
P\big(\hat{y}^\text{Data}\big|h_\text{CLS}^\text{Data}\big) &= \text{Softmax}\left(\text{Linear-Head}^\text{Data}\left(h_\text{CLS}^\text{Data}\right) \right),\\
\mathcal{L}_\text{Data} &= -\log P\big(\hat{y}^\text{Data}=y\big|h_\text{CLS}^\text{Data}\big),
\end{align}
where $\text{Linear-Head}^\text{Data}$ is the linear classification head for Data Encoder and the Cross-entropy loss is utilized.

We continue to finetune all the Transformer parameters during multimodal training. Prompt methods adopted in News Encoder are not reserved here because the 6-layer Transformer (1.7M parameters) is much smaller in size than the RoBERTa model (335M parameters).

\subsection{Fusion of News Encoder and Data Encoder}
\label{sec:fusion}

News Encoder and Data Encoder provide us with the text feature $h_\text{CLS}^\text{News}$ and the time series feature $h_\text{CLS}^\text{Data}$ respectively. We build our multimodal fusion block on top of those features. We first linearly project $h_\text{CLS}^\text{News}$,$h_\text{CLS}^\text{Data}$ into $v^\text{News}$, $v^\text{Data}$ which lie in a common embedding space $\mathbb{R}^{d_\text{Align}}$:
\begin{equation}
   v^\text{News}, v^\text{Data}=\text{Linear}^\text{News}(h_\text{CLS}^\text{News}), \text{Linear}^\text{Data}(h_\text{CLS}^\text{Data}).
\end{equation}

$v^\text{News}$, $v^\text{Data}$ receive the supervision signal from the contrastive alignment, which will be discussed in the next section, to learn multimodal representations. 

The fusion prediction $\hat{y}^\text{Fusion}$ and fusion loss $\mathcal{L}_\text{Fusion}$ is generated from $v^\text{News}$, $v^\text{Data}$ with a fusion process $\text{Fusion}(v^\text{News}, v^\text{Data})$. Similarly, the cross-entropy loss is also adopted:
\begin{align}
P\big(\hat{y}^\text{Fusion}\big|v^\text{News}, v^\text{Data}\big) &= \text{Softmax}\left(\text{Fusion}(v^\text{News}, v^\text{Data})\right),\\
\mathcal{L}_\text{Fusion} &= -\log P\big(\hat{y}^\text{Fusion}\big|v^\text{News}, v^\text{Data}\big).
\end{align}

We have implemented several fusion methods in our experiments and we find that the linear fusion function achieves the best result, which is shown in our analysis. The linear fusion function is designed as:
\begin{align}
\label{eq:linear_fusion}
\text{Fusion}_\text{Linear}(v^\text{News}, v^\text{Data})=W^\text{News}v^\text{News}+W^\text{Data}v^\text{Data}+b.
\end{align}
where $W^\text{News}$,$W^\text{Data}\in \mathbb{R}^{2 \times d_\text{Align}}$, $b \in \mathbb{R}^2$.

\subsection{Cross-Modality Contrastive Alignment}
\label{sec:alignment}

Contrastive learning is a widely used technique in the multi-modal area. Methods like CLIP~\citep{clip} and ALPRO~\citep{alpro} use cross-modality contrastive losses to align representations in different embedding spaces. In our model, we implement the cross-modality contrastive alignment between overnight news and historical trading data with the alignment loss. 

Only matched pairs in a batch are considered to be positive pairs. Given a batch of pair $V^\text{News}=[v_1^\text{News},v_2^\text{News},\cdots,v_B^\text{News}]$,$V^\text{Data}=[v_1^\text{Data},v_2^\text{Data},\cdots,v_B^\text{Data}] \in \mathbb{R}^{B\times d_\text{Align}}$ and $B$ represents the batch size, we define the similarity using dot-product following~\cite{clip, align, alpro}:
\begin{equation}
    \text{Sim}(i,j)=v_i^\text{News} \cdot v_j^\text{Data},
\end{equation}
and the two symmetric News-to-Data (N2D), Data-to-News (D2N) alignment losses are subsequently defined as:
\begin{align}
    \mathcal{L}_\text{Align}^\text{N2D} =& - \frac{1}{B} \sum_{i=1}^{B}\log \frac{\exp\big(\nicefrac{\text{Sim}(i,i)}{\tau}\big)}{\sum\limits_{j=1}^B \exp\big(\nicefrac{\text{Sim}(i,j)}{\tau}\big)},\\
    \mathcal{L}_\text{Align}^\text{D2N} =& - \frac{1}{B} \sum_{j=1}^{B}\log \frac{\exp\big(\nicefrac{\text{Sim}(j,j)}{\tau}\big)}{\sum\limits_{i=1}^B \exp\big(\nicefrac{\text{Sim}(i,j)}{\tau}\big)}.
\end{align}

Here $\tau$ is the temperature parameter. The total cross-modality contrastive alignment loss is:
\begin{align}
\mathcal{L}_\text{Align}=\mathcal{L}_\text{Align}^\text{N2D}+\mathcal{L}_\text{Align}^\text{D2N}.
\end{align}

Note that previous works use the cross-modality contrastive loss to boost performance on retrieval tasks such as image classifications, and some even report harm on other tasks~\citep{alpro}. 

However, in our model, $\mathcal{L}_\text{Align}$ is designed to improve representation learning in different modalities, and the defined similarity $\text{Sim}$ is not used for inference. 

In addition, embedding spaces of text and historical data can be significantly different because they are not so strongly connected as typical settings in image-text or video-text scenarios. Deep connections between text-data modalities can cause models to learn incorrect relations and fall into the trap of overfitting as shown in our analysis experiments. We find that as the relatively simple cross-modality contrastive alignment loss is applied to the output of the encoders, it encourages multimodal fusion and alignment and does not harm the structures of the large models.

\subsection{Training Objectives}
\label{sec:training_objective}
We combine the aforementioned four different losses for our final training objectives. Two losses $\mathcal{L}_\text{News}$,$ \mathcal{L}_\text{Data}$ are derived from unimodal encoders solely based on $h_\text{CLS}^\text{News}$,$h_\text{CLS}^\text{Data}$ respectively. Moreover, we leverage the fusion of $v^\text{News}$,$v^\text{Data}$ to construct a multimodal prediction for the target labels' distribution. The final training objective is a weighted sum:
\begin{equation}
\mathcal{L}=\lambda_\text{N}\mathcal{L}_\text{News}+\lambda_\text{D}\mathcal{L}_\text{Data}+\lambda_\text{F}\mathcal{L}_\text{Fusion}+\lambda_\text{A}\mathcal{L}_\text{Align}.
\end{equation}
\subsection{Inference}
\label{sec:inference}

Our prediction results also consist of three elements as mentioned before. We use the equally weighted average to produce the ensembled predicted probabilities:
\begin{equation}
    P\left(\hat{y}\right)=\frac{ P\left(\hat{y}^\text{News}\right)+
    P\left(\hat{y}^\text{Data}\right)+
    P\left(\hat{y}^\text{Fusion}\right)}{3}, 
\end{equation}
and the final predicted label is generated by:
\begin{equation}
\hat y=\argmax\limits_y P(y|X,W),
\end{equation}
where $X$ denotes the input historical trading data and $W$ denotes the stock-related financial news.

Note that our methods can still be functional if one of the two modal inputs is missing. This can be done by simply disabling the corresponding encoder, making our model more robust toward different input situations.

\section{Experiments}

In this section, we first introduce the datasets, then we describe the baseline algorithms, detailed settings, and the experimental results.

\subsection{Datasets and Data Processing}
The historical trading data is extracted from TOPIX500, which is comprised of the 500 most liquid and highly market-capitalized stocks in Tokyo Stock. We split the dataset chronologically to avoid information leakage. The data from Jan. $1^\text{st}$, 2013 to Dec. $31^\text{st}$, 2017 is used as the training set, and 
Jan. $1^\text{st}$, 2018 to Apr. $30^\text{th}$, 2018 as the development set, May $1^\text{st}$, 2018 to Sept. $30^\text{th}$, 2018 as the test set. During data processing, we drop the data point where there is a missing entry. The overnight news is collected from Reuters Financial News\footnote{https://github.com/liweitj47/overnight-stock-movement-prediction}. Following~\citep{chen2019incorporating,li2021modeling,zhao2021long}, the data is filtered with RIC labels provided by Reuters, which are the possible stocks that may be influenced by the news.

\begin{table}[t]
\caption{Dataset statistics for stock movement prediction.}
\vskip -0.05 in
\label{tab:multimodal_dataset}
\centering
\begin{tabular}{@{}lccc}
\toprule
 \bf Split & \bf Train &  \bf Dev  & \bf Test\\ 
 \midrule
\bf Samples & 8,483 & 687 & 938\\
 \bottomrule
\end{tabular}
\vskip -0.1 in
\end{table}


\begin{table}[t]
\caption{Dataset statistics for Data Encoder pre-training.}
\vskip -0.05 in
\label{tab:pre-train dataset}
\centering
\begin{tabular}{@{}lccc}
\toprule
 \bf Split & \bf Train &  \bf Dev  & \bf Test\\ 
 \midrule
\bf Samples & 74,950 & 2,214 & 4,072\\
 \bottomrule
\end{tabular}
\vskip -0.1 in
\end{table}

\begin{table*}[t]
\caption{Main results. Existing methods can mainly be classified into statistical, unimodal, fusion-only, and ensemble-only methods, while our proposed ProMUSE achieves state-of-the-art performance.}
\vskip -0.05 in
\label{tab:main_results}
\centering
\begin{tabular}{@{}llllc}
\toprule
\multicolumn{2}{c}{\bf Method} & {\bf News Encoder} & {\bf Data Encoder} & {\bf ACC} \\ 
\midrule
\multirow{2}{*}{\textbf{Statistical}} & Random & - & - & 50.00 \\
 & 20-day EMA & - & - & 59.38 \\
 \midrule
\multirow{5}{*}{\shortstack{\bf Unimodal \\ (News)}} & \multirow{2}{*}{\shortstack{Training\\ from scratch}} & \multirow{2}{*}{\shortstack{Transformer}} & \multirow{2}{*}{-} & \multirow{2}{*}{58.26}\\
& \\
\cmidrule{2-5}
& Zero-shot & Financial-RoBERTa & - & 48.82\\
& Fine-tuning & Financial-RoBERTa & -& 62.19 \\
& Prompt & Financial-RoBERTa & - & \textbf{63.75} \\
 \midrule
\multirow{5}{*}{\shortstack{\bf Unimodal \\ (Data)}} & \multirow{3}{*}{\shortstack{Training\\ from scratch}} & - & Linear & 58.93 \\
& & - & LSTM & 54.78\\
&  & - & Transformer & 66.07 \\
\cmidrule{2-5}
& Zero-shot & - & Pre-trained Transformer &  64.71 \\
 & Fine-tuning & - & Pre-trained Transformer &\textbf{68.44} \\
 \midrule
\multirow{4}{*}{\bf Fusion-Only} & \multirow{3}{*}{\shortstack{Training\\ from scratch}} & Transformer & Linear &  61.38 \\
&  & Transformer & LSTM & 57.70 \\
&  & Transformer & Transformer & 60.16 \\
\cmidrule{2-5}
& Prompt & Financial-RoBERTa & Pre-trained Transformer & \textbf{72.47} \\
 \midrule
\multirow{5}{*}{\bf Ensemble-Only} & \multirow{4}{*}{\shortstack{Training\\ from scratch}}  & Transformer & Linear & 58.48\\
&  & Transformer & LSTM & 57.92 \\
&  & Transformer & Transformer & 60.60\\
&  & Transformer & Linear+LSTM+Transformer & 61.94\\
\cmidrule{2-5}
& Prompt (Average) & Financial-RoBERTa & Pre-trained Transformer& \textbf{72.28} \\
 \midrule
{\bf ProMUSE} & \textbf{Prompt} & Financial-RoBERTa & Pre-trained Transformer&\textbf{73.56$^\star$} \\
\bottomrule
\end{tabular}

\end{table*}

In this paper, we only select the data in which both the overnight news and the historical trading data are available for stock movement prediction.  In addition, we filter the data where the volume movement is not significant enough\citep{zhao2021long}, alleviating the effect of randomness and minor, irrelevant news:
\begin{align}
    \sigma^v=\sqrt{\frac{1}{20}\sum_{i=1}^{20} \left(x_{i,1}^v-\bar{v}\right)^2},\quad s^v=\frac{x_{21,1}^v-\bar{v}}{\sigma^v}.
\end{align}

If $\abs{s^v} \leq 0.5$, we consider its volume movement insignificant and remove it from the dataset. The statistical information for the final dataset can be found in Table~\ref{tab:multimodal_dataset}.

For pre-training our Data Encoder, we use the same split as the stock movement prediction dataset. The dataset statistics are demonstrated in Table ~\ref{tab:pre-train dataset}.

\subsection{Baselines}
In this section, we introduce the algorithms of the baseline models, including traditional statistical methods, unimodal models, fusion methods, and ensemble methods.

\subsubsection{Statistical Methods}
\begin{itemize}
    \item \textbf{Random:} Randomly predict the label $\hat{y}\sim B(1,1/2)$.
    \item \textbf{Exponential Moving Average (EMA):} In this task, the EMA series are defined as:
    \begin{equation}
        \text{EMA}_n=\frac{1}{21}\left(2 \text{EMA}_{n-1}+ 19x_{n,1}^v
    \right),
    \end{equation}
    where $\text{EMA}_1$ is initialized as $x_{1,1}^v$, and we use $\text{EMA}_{20}$ as a prediction of $x_{21,1}^v$. The prediction is $\hat{y}=\mathbb{I}\left(\text{EMA}_{20}>\bar{v}\right)$.
    
\end{itemize}

\subsubsection{Models Training from Scratch}
We try to train a six-layer Transformer from scratch with the hidden size set to 200 to replace the pre-trained Financial-RoBERTa in the News Encoder. With regard to the Data Encoder, we try multiple structures including linear, one-layer LSTM~\citep{lstm}, and a six-layer Transformer. The detailed structure is similar to \citet{zhang2023asat}. We also explore different fusion or ensembling paradigms for model training from scratch.

\subsubsection{Unimodal Models}
For unimodal models, we use only one of the encoders. As a result, the final training objective and predicted probability distribution degrades into only $\mathcal{L}_\text{News}$, $P\left(\hat{y}^\text{News}\right)$ or $\mathcal{L}_\text{Data}$, $P\left(\hat{y}^\text{Data}\right)$.

\subsubsection{Fusion Methods}

\begin{itemize}
    \item \textbf{Attention:} Using attention mechanism to produce the weights for fusion. $w_1$,$w_2 \in \mathbb{R}^{d_\text{Align}}$ is derived linearly from $v^\text{News}$ and $v^\text{Data}$:
    \begin{equation}
        w_1,w_2= \text{Softmax}\left(\text{Linear}([v^\text{News},v^\text{Data}])\right),
    \end{equation}
    and the fusion function is:
    \begin{equation}
    \label{eq:attention_fusion}
        \text{Fusion}_\text{Attention}=\text{Linear}\left(w_1 \odot v^\text{News}+ w_2 \odot v^\text{Data}\right).
    \end{equation}
    Here $\odot$ represents element-wise multiplication.
    
    \item \textbf{Transformer:} Stack a multimodal Transformer encoder on top of both encoders. It takes their last-layer hidden states as input after transforming to the same dimension and uses a linear classification head for prediction. 
\end{itemize}

\subsubsection{Ensemble Methods}
In this section, we introduce the ensemble methods based on the unimodal models' predictions, $P(\hat{y}^\text{News})$ and $P(\hat{y}^\text{Data})$.

\begin{itemize}
    \item \textbf{Learnable Weights:} We set $w \in \mathbb{R}$ as a learnable parameter, and initialize it to 1. The formulation is $P(\hat{y}^\text{Learnable})= \\ \text{Sigmoid}(w)P(\hat{y}^\text{News})+(1-\text{Sigmoid}(w))P(\hat{y}^\text{Data})$.

    \item \textbf{Predicted Weights:} Use a linear head to predict the weights between two modalities: 
    \begin{equation}
        w_1,w_2=\text{Softmax}(\text{Linear}([h_\text{CLS}^\text{News},h_\text{CLS}^\text{Data}])),
    \end{equation}
    and then $P(\hat{y}^\text{Predicted})=w_1 P(\hat{y}^\text{News})+w_2 P(\hat{y}^\text{Data})$.
    
    \item \textbf{Normalized:} Normalize the unimodal results to standardized Gaussian distribution:  $P(\hat{y}^\text{Norm})=\text{Norm}(P(\hat{y}^\text{News}))+ \\ \text{Norm}(P(\hat{y}^\text{Data}))+0.5$, where $\text{Norm}(p)=(p-\mu)/\sigma$ and hyperparameters $\mu$ and $\sigma$ for corresponding probabilities $P(\hat{y}^\text{News})$ or $P(\hat{y}^\text{Data})$ are calculated on the development set.
    
\end{itemize}

\subsection{Settings and Hyperparameters}
We train every model in our experiments for 40 epochs and report the test accuracy as the result. The checkpoints with the best accuracy on the development set are selected for the report. We repeat every experiment 4 times and report the average results.

In our main experiments, we adopt the AdamW optimizer~\citep{loshchilov2017decoupled}, using the learning rate as 1e-5 and the weight decay factor as 1e-3. The batch size $B$ is chosen as 32. The weights for each loss are set as $\lambda_N=\lambda_D=\lambda_F=1$ and $\lambda_\text{Align}=0.1$. $d_\text{Align}$ is set to 200. We use the log value of the stock historical trading data $X$ as the input of the Data Encoder.

For the pre-training of our Data Encoder, we run 100 epochs and save the checkpoint with the lowest development loss for the main experiments. We also use the AdamW optimizer with the learning rate as 1e-5 and the weight decay factor as 1e-3. Additional Exponential decay is used with a factor of 0.95. Pre-training batch size is set for 64.

\subsection{Experimental Results}
Our main experimental results can be found in Table~\ref{tab:main_results}. 

\subsubsection{Training from Scratch} 

We adopt a six-layer Transformer as an alternative to the Financial-RoBERTa model in the News Encoder. In the case of the Data Encoder, we explore linear, LSTM, and Transformer models as substitutes for the pre-trained Transformer. The outcomes reveal that the available data volume is insufficient for training a News Encoder from scratch, while our method successfully utilizes the knowledge of the pre-trained Financial RoBERTa. Furthermore, the linear and LSTM structures yield inferior performance compared to Transformer. The benefits of employing a pre-trained Transformer model are also evident.

\subsubsection{Unimodal Models}
Unimodal models perform significantly worse than multimodal methods, as only the information of a single modality is fed to them, and both modalities are important in this task. Besides, we find that prompting the Financial-RoBERTa can more effectively leverage its knowledge and performs better than fine-tuning. Moreover, modeling historical trading data is easier than financial news, thus data-only methods can achieve better performance than new-only models.

\subsubsection{Fusion-Only and Ensemble-Only Methods}

Direct multimodal fusion or ensemble can achieve improvement compared with unimodal methods. However, fusion-only methods without unimodal supervision may inflict damage on unimodal representations, and ensemble-only methods lack multimodal connections. Our method settles these problems by constructing fusion and cross-modality contrastive alignment as well as retaining unimodal predictions to help representation learning.

\subsubsection{Effectiveness of ProMUSE}

As Table~\ref{tab:main_results} shows, ProMUSE can achieve the best performance. We successfully surpass the widely-used traditional EMA baseline in the financial area and exhibit our advantages against the models training from scratch, unimodal models, fusion-only methods, and ensemble-only methods.

\section{Analysis}
In this section, we first conduct the ablation study to prove the effectiveness of each module in our model. Then discuss the more detailed parts, the exploration of the tuning paradigm, fusion models, ensemble algorithms, and prompt learning methods. Finally, we present that our method helps the representation learning process.

\subsection{Ablation Study}

\begin{table}[!t]
\caption{Ablation study results.}
\vskip -0.05 in
\renewcommand\arraystretch{0.9}
\label{tab:ablation_results}
\setlength{\tabcolsep}{1pt}
\centering
\begin{tabular}{@{}lcccccccc}
\toprule
\scriptsize \bf Method &  \scriptsize $\mathcal{L}^\text{News}$ & \scriptsize $\mathcal{L}^\text{Data}$  & \scriptsize$\mathcal{L}^\text{Fusion}$ & \scriptsize$\mathcal{L}^\text{Align}$ & \scriptsize $P(\hat{y}^\text{News})$ & \scriptsize$P(\hat{y}^\text{Data})$ & \scriptsize$P(\hat{y}^\text{Fusion})$ & {\bf ACC} \\ 
\midrule
\scriptsize Fusion-Only & \checkmark & \checkmark & \checkmark & $\times$ & $\times$ & $\times$ & \checkmark &72.47 \\
\scriptsize Ensemble-Only & \checkmark & \checkmark & $\times$ & $\times$ & \checkmark & \checkmark & $\times$ &72.28 \\
\midrule
\scriptsize w/o News Head & $\times$ & \checkmark & \checkmark & \checkmark & $\times$ & \checkmark & \checkmark &72.15\\
\scriptsize w/o Data Head & \checkmark & $\times$ & \checkmark & \checkmark & \checkmark & $\times$ & \checkmark &69.91 \\
\scriptsize w/o News/Data & $\times$ & $\times$ & \checkmark & \checkmark & $\times$ & $\times$ & \checkmark & 70.55 \\
\midrule
\scriptsize w/o Fusion & \checkmark & \checkmark & $\times$ & \checkmark & \checkmark  & \checkmark & $\times$ &67.80 \\
\scriptsize w/o Alignment & \checkmark & \checkmark & \checkmark & $\times$ & \checkmark  & \checkmark & \checkmark & 70.52 \\
\scriptsize w/o Ensemble & \checkmark & \checkmark & \checkmark & \checkmark & $\times$ & $\times$  & \checkmark & 73.35 \\
\midrule
\textbf{ProMUSE} & \checkmark & \checkmark & \checkmark & \checkmark & \checkmark & \checkmark &\checkmark & \textbf{73.56} \\
\bottomrule
\end{tabular}
\end{table}

\begin{table}[!t]
\caption{Results of different fusion models. The models below are fusion-only models without unimodal supervision. The linear fusion outperforms other sophisticated fusions.}
\vskip -0.05 in
\label{tab:fusion}
\centering
\begin{tabular}{@{}lc}
\toprule
\bf Fusion Model  &{\bf ACC} \\ 
\midrule
Attention  &  67.00 \\
One-layer Transformer  &  65.43 \\
Six-layer Transformer  &  69.11 \\
Fusing data into the last layer in News Encoder &  63.81 \\
Fusing data to every layer in News Encoder & 63.86 \\
\midrule
\textbf{Linear}  &  \textbf{72.47}  \\
\bottomrule
\end{tabular}
\vskip -0.1 in
\end{table}

\subsubsection{Effectiveness of Different Losses and Prediction Heads}

We conduct various ablation experiments to test the effectiveness of each component in our proposed ProMUSE model shown in Table~\ref{tab:ablation_results}. 

The results prove that the combination of fusion, ensemble, and cross-modality contrastive alignment reaches the best performance. The ensemble-only method can cause degradation where accuracy falls from 73.56 to 72.28 as cross-modality interaction is lost. The fusion-only setting also triggers a loss in accuracy to 72.47 due to a heavier influence on unimodal representations without $\mathcal{L}^\text{Align}$. Dropping unimodal heads will severely damage the performance as both modalities are indispensable. The removal of the fusion or alignment process also breaks the connections between modalities.

Note that during the inference stage, predicting $P(\hat{y}^\text{Data})$ and $P(\hat{y}^\text{News})$ does not need much extra computation as the calculation of $P(\hat{y}^\text{Fusion})$ also needs the calculation of two heads. Predicting them for ensemble can get an improvement from 73.35 to 73.56.

\subsubsection{Effectiveness of Fusion Models}

In this section, we further explore different fusion models in Table~\ref{tab:fusion}. The linear fusion method which is adopted in our proposal outperforms all other variants. Here Attention is described in equation~\ref{eq:attention_fusion}, One-layer/Six-layer Transformer is stacked on the two encoders as discussed in the baseline section. Fusing data into the News Encoder methods use $h_\text{CLS}^\text{Data}$ to generate the continuous prompts for the News Encoder.

The reason is that the two encoders are pre-trained and will not easily overfit into the knowledge of the small dataset, while the complicated fusion methods such as introducing another Transformer model are trained from scratch, which create additional parameters and can lead to incorrect connections and overfitting.

\subsubsection{Effectiveness of Ensemble Algorithms}

\begin{figure}[!t]
    \centering
    \includegraphics[width=0.8\linewidth]{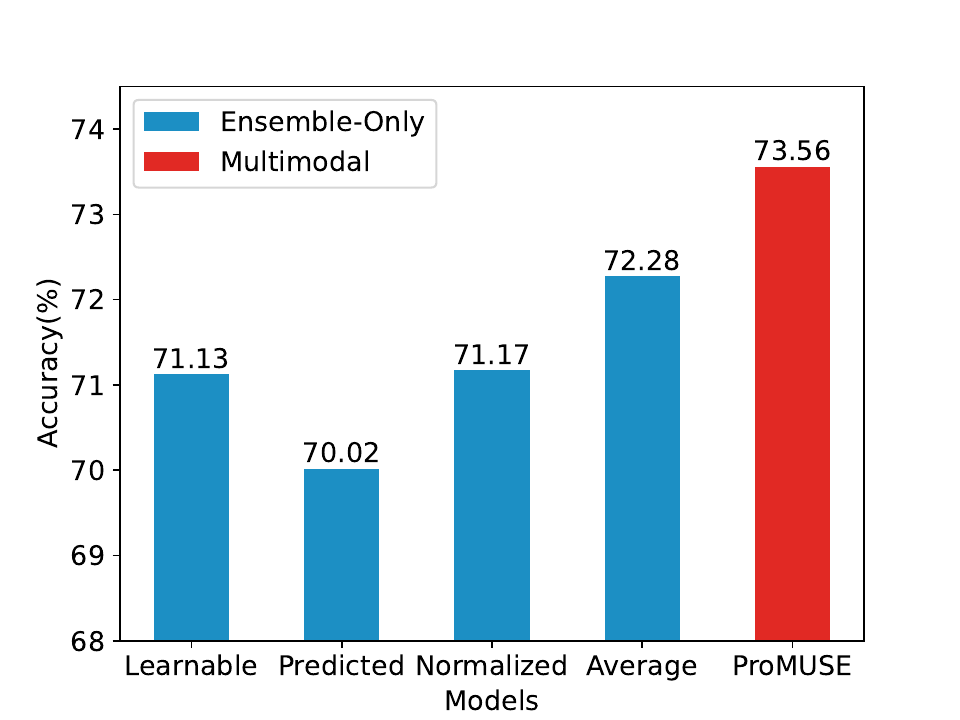}
    \vskip -0.05 in
    \caption{Results of different ensemble algorithms. The average of the two unimodal outputs exceeds other algorithms.}
    \label{fig:ensemble}
    \vskip -0.1 in
\end{figure}

We test the effectiveness of different ensemble algorithms in Figure~\ref{fig:ensemble}. In our experiments on ensemble-only models, we find that an average of news prediction and data prediction perform best. In accordance with our previous analysis of fusion models, complicated ensemble algorithms also tend to deteriorate the performance, because new parameters involved in ensembling are estimated on a small dataset, which are difficult to generalize and may cause overfitting. Therefore we choose to calculate the algorithmic mean of $P(\hat{y}^\text{News})$, $P(\hat{y}^\text{Data})$ and $P(\hat{y}^\text{Fusion})$ for inference in our model.

\subsection{Why We Utilize P-Tuning v2 Paradigm}

\begin{table}[!t]
\caption{Results of different prompt paradigms. P-Tuning v2 outperforms hard prompt, soft prompt, and fine-tuning.}
\vskip -0.05 in
\renewcommand\arraystretch{0.9}
\label{tab:Prompt}
\centering
\begin{tabular}{@{}llc}
\toprule
\bf Model & Tuning Method &{\bf ACC} \\ 
\midrule
\multirow{4}{*}{\shortstack{News Encoder}} & Hard Prompt &   48.82\\
 & Soft Prompt &  56.08 \\
 & Fine-tuning & 62.19  \\
\cmidrule{2-3}
 & \textbf{P-Tuning v2} & \textbf{63.75}  \\
\midrule
\multirow{2}{*}{\shortstack{Data Encoder}} &  Zero-Shot & 64.71  \\
\cmidrule{2-3}
 & \textbf{Fine-tuning} &  \textbf{68.44} \\
\midrule
\multirow{2}{*}{\textbf{ProMUSE}} & Fine-tuning  &  66.79 \\
\cmidrule{2-3}
 & \textbf{Prompt} & \textbf{73.56$^\star$}  \\
\bottomrule
\end{tabular}
\vskip -0.1 in
\end{table}

In search of the proper paradigms for our News Encoder and Data Encoder, we conduct the experiments in Table~\ref{tab:Prompt}. The template we use for the hard prompt in our experiments is \textit{News: $\{w_1,w_2,\cdots,w_n\}$. The volume will go up/down.} Soft prompt here represents implementing continuous prompts in the embedding layer, which are similar to Prefix-Tuning~\citep{prefix-tuning} and P-Tuning~\citep{ptuning}.

We find that P-Tuning v2 best suits the News Encoder and achieve an accuracy of 63.75, which inserts continuous prompts in every layer due to more trainable parameters and deeper layers. 
Fine-tuning the Data Encoder can get great improvement against zero-shot setting, but fine-tuning the large Financial-RoBERTa model with limited data can easily get into the overfitting problem and cause sub-optimal, with the accuracy dropping to 66.79.

\begin{figure}
    \centering
    \includegraphics[width=0.8\linewidth]{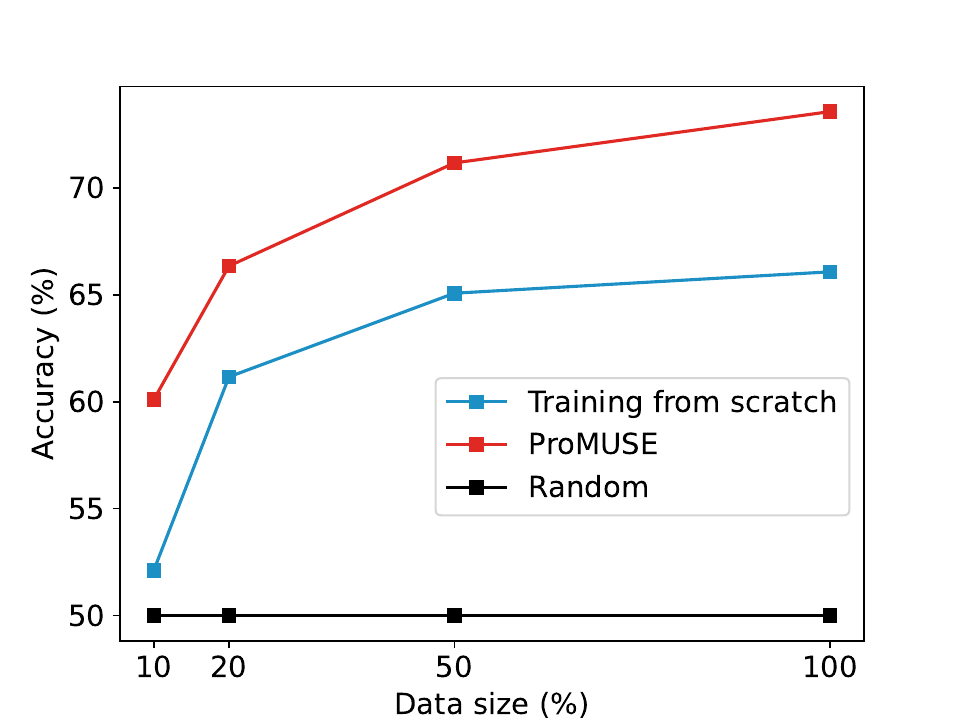}
    \vskip -0.05 in
    \caption{Results of varying data size. ProMUSE outperforms methods training from scratch under lower resources.}
    \label{fig:prompt}
    \vskip -0.05 in
\end{figure}

\subsection{ProMUSE Works under Lower Resources}

We further analyze the effect of our prompt-based multimodal model by varying the size of the training data. The development set and test set are kept unchanged, and we train all models for 80 epochs to compensate for the reduction of data volume. 

Here we compare the performance of ProMUSE with a six-layer transformer which is trained from scratch and only receives historical trading data input because it performs best in our previous experiments. We drop News Encoder as the data volume continues to shrink and a high-quality encoder cannot be obtained under these circumstances. The results are shown in Figure~\ref{fig:prompt}. We find that ProMUSE can achieve significant improvement over training from scratch in all settings, proving that exploiting the universal knowledge of pre-trained language models through prompt learning is essential in this task.

\subsection{ProMUSE Helps Representation Learning}

\begin{table}[!t]
\caption{Analysis of the damage of multimodal learning to unimodal representations. Existing methods corrupt unimodal representations, while ProMUSE can mitigate the damage of multimodal learning to unimodal representations.}
\vskip -0.05 in
\label{tab:alignment}
\centering
\begin{tabular}{@{}lccc}
\toprule
\bf Model & News-Only & Data-Only & Multi-Modal \\ 
\midrule
\bf News Encoder & \textbf{63.75} & - & 63.75\\ 
\bf Data Encoder & - & \textbf{68.44} & 68.44\\ 
\midrule
\textbf{w/o Alignment}  & 58.32&  61.09 & 70.52 \\
\textbf{w/o Fusion} &  62.60 & 63.65 & 67.80 \\
\midrule
{\textbf{ProMUSE}} & 62.69 & 67.38 & \textbf{73.56}  \\ 
\bottomrule
\end{tabular}
\vskip -0.05 in
\end{table}

In this section, we discuss why our proposal can achieve improvement. As Table~\ref{tab:alignment} shows, the introduction of multi-modal learning can get general gains compared to unimodal models. However, simultaneously training the two modalities can harm the representation learning of the individual encoders, as we see a decrease in accuracy for news-only or data-only input scenarios. 

Our model uses the cross-modality contrastive alignment and multi-modal fusion to better alleviate the damage caused in the multi-modal learning process, providing a strong constraint and regularization for the unimodal encoders to best avoid degradation.

\section{Related Work}

\subsection{Pre-trained Language Models}
Since Tranformer~\citep{transformer} shows great success in the natural language processing (NLP) area, various pre-trained language models are proposed and achieve state-of-the-art performance in numerous NLP tasks. Models such as 
and GPT-3~\citep{gpt3} exploit Transformer decoder structure to construct unidirectional autoregressive language models. Bidirectional BERT-like models~\citep{bert,roberta,albert} are basically based on Transformer encoders. T5~\citep{t5}, BART~\citep{bart}, and Flan-T5~\citep{flan-t5} choose to adopt the encoder-decoder framework. They are trained on corresponding pre-training tasks and large unlabelled corpora, and models become increasingly large in size~\citep{opt,palm} to pursue better performance. Recently, Reinforcement Learning from Human Feedback techniques are applied to ChatGPT and GPT-4~\citep{gpt4}, which gain outstanding performance.

\subsection{Multimodal Learning}
Existing multimodal learning methods mainly focus on image-text and video-text tasks. UNITER~\citep{uniter} learns joint contextualized representations for both text and image through pretraining. ViLBERT~\cite{vilbert} extends the BERT model by the co-attentional Transformer layers for learning task-agnostic representations of image content and natural language. ALIGN~\citep{align} and CLIP~\citep{clip} construct the dual-encoder architecture to align visual and textual representations using image-text contrastive learning, and the cross-modality contrastive loss has become an important component in many models. BLIP-2~\citep{blip-2} uses a lightweight querying Transformer to learn from frozen image encoders and large language models.

\subsection{Stock Movement Prediction}
Stock movement prediction is a key research direction in the finance area. \citet{xu2018stock} present a deep generative model to jointly learn from tweet text and price signals, and use recurrent latent variables to process stochasticity. \citet{li2021modeling} design an LSTM-RGCN model for learning overnight news and the correlation between stocks. \citet{chen2022stock} set up a dual-process meta-learning method to mine general patterns and stock-specific knowledge. \citet{xie2023wall} analyze the zero-shot ability of ChatGPT in multimodal stock movement prediction.

\section{Conclusion}
In this paper, we present ProMUSE, a prompt-based multimodal stock volume movement prediction model, to fully exploit the textual information in financial news and the potential of pre-trained language models with limited data. We use the News Encoder and Data Encoder to process the overnight financial news and historical trading data respectively, and use a fusion model to generate multimodal output. Unimodal supervision, multimodal supervision, and cross-modality contrastive alignment are used for training, while unimodal and multimodal predictions constitute the final inference result. Extensive experiments show that our method significantly outperforms various baselines. Comprehensive analysis testifies to the effectiveness of different modules. Moreover, ProMUSE can help mitigate the harm to representation learning during joint training of textual and time series modalities.


\clearpage
\bibliographystyle{ACM-Reference-Format}
\bibliography{sample-base}

\end{document}